\documentclass[a4paper,fleqn]{cas-dc}

\usepackage[numbers, sort&compress, longnamesfirst]{natbib}
\usepackage[T1]{fontenc}
\usepackage{algorithmic}
\usepackage{algorithm}
\usepackage{subfigure}
\usepackage{amsmath}
\usepackage{multirow}
\usepackage[normalem]{ulem}
\useunder{\uline}{\ul}{}
\usepackage{amsmath,amssymb,amsfonts}
\usepackage{graphicx}

\def\tsc#1{\csdef{#1}{\textsc{\lowercase{#1}}\xspace}}
\tsc{WGM}
\tsc{QE}
\tsc{EP}
\tsc{PMS}
\tsc{BEC}
\tsc{DE}
\begin{document}
\let\WriteBookmarks\relax
\def\floatpagepagefraction{1}
\def\textpagefraction{.001}

\shorttitle{Robust Adversarial Attacks Detection for Deep Learning based Relative Pose Estimation for Space Rendezvous}
\shortauthors{Z. Wang et~al.}

\title [mode = title]{Robust Adversarial Attacks Detection for Deep Learning based Relative Pose Estimation for Space Rendezvous}    
%\tnotetext[1]{This paper is part of the results of the project funded by the European Space Agency. Project ID: ESA AO/2-1856/22/NL/GLC/ov.}

\let\printorcid\relax
\author[1]{Ziwei Wang}
\cormark[1]
\ead{ziwei.wang.3@city.ac.uk}
        
\author[1]{Nabil Aouf}
%\ead{nabil.aouf@city.ac.uk}

\author[2]{Jose Pizarro}
%\ead{jose.pizarro@esa.int}

\author[2]{Christophe Honvault}
%\ead{christophe.honvault@esa.int}

\affiliation[1]{organization={Department of Engineering},
            addressline={City, University of London, Northampton Square}, 
            city={London},
            postcode={EC1V 0HB}, 
            country={United Kingdom}}
            
\affiliation[2]{organization={European Space Agency},
            addressline={European Space Research \& Technology Centre, Keplerlaan 1, PO Box 299}, 
            city={Noordwijk},
            postcode={2200 AG}, 
            country={The Netherlands}}

\cortext[1]{Corresponding author}

\begin{abstract}
Research on developing deep learning techniques for autonomous spacecraft relative navigation challenges is continuously growing in recent years. Adopting those techniques offers enhanced performance. However, such approaches also introduce heightened apprehensions regarding the trustability and security of such deep learning methods through their susceptibility to adversarial attacks. In this work, we propose a novel approach for adversarial attack detection for deep neural network-based relative pose estimation schemes based on the explainability concept. We develop for an orbital rendezvous scenario an innovative relative pose estimation technique adopting our proposed Convolutional Neural Network (CNN), which takes an image from the chaser's onboard camera and outputs accurately the target's relative position and rotation. We perturb seamlessly the input images using adversarial attacks that are generated by the Fast Gradient Sign Method (FGSM). The adversarial attack detector is then built based on a Long Short Term Memory (LSTM) network which takes the explainability measure namely SHapley Value from the CNN-based pose estimator and flags the detection of adversarial attacks when acting. Simulation results show that the proposed adversarial attack detector achieves a detection accuracy of 99.21\%. Both the deep relative pose estimator and adversarial attack detector are then tested on real data captured from our laboratory-designed setup. The experimental results from our laboratory-designed setup demonstrate that the proposed adversarial attack detector achieves an average detection accuracy of 96.29\%. 
\end{abstract}

% \includegraphics{figs/grabs.pdf}

%\item Research highlights item 1
%\item Research highlights item 2
%\item Research highlights item 3

% Keywords
\begin{keywords}
 Pose Estimation \sep Adversarial Attack \sep Adversarial Attack Detection  \sep Explainable Artificial Intelligence \sep Deep Learning
\end{keywords}

\maketitle

\section{Introduction}

The growth of deep learning-based techniques has drawn increasing attention in various domains of application, such as image processing, speech recognition, and many other challenging Artificial Intelligence (AI) based tasks~\cite{guo2016deep}. Vision-based autonomous orbital space rendezvous, ~\cite{wie2014attitude}, is an application for which adopting deep learning approaches to spacecraft position and attitude estimation is continuously gaining interest within the research community and the space agencies~\cite{song2022deep,kisantal2020satellite}. 

The state-of-the-art achievements in deep learning (DL) research demonstrate that the Convolutional Neural Networks (CNNs) have successfully gained outstanding performance in computer vision applications, such as object detection and target localisation~\cite{ren2017faster, redmon2018yolov3,cebollada2022development}. Determining the pose of a spacecraft's relative state by processing input images is typically achieved through the six Degree-of-Freedom (6 DOF) pose estimation of the target object frame relative to the camera (onboard the spacecraft) frame. These vision-based pose estimation methods are traditionally computed by matching relative features on images captured by the camera to relative locations in the target frame. Different from the traditional approaches, the CNNs can be trained to detect features from raw image data and estimate the relative pose by regressing the position and attitude, without the need for manual feature engineering which is often required in traditional computer vision methods. The advantages of CNN-based pose estimation approaches are that they can potentially lead to better performance in complex orbital scenarios and more robustness to variations in lighting, viewpoint, and cluttered background. 

Recent achievements in CNN-based pose estimation demonstrate outstanding accuracy performance~\cite{phisannupawong2020vision,oestreich2020orbit,rondao2022chinet}. However, The vulnerability of such deep learning scheme can be questionable. Indeed, minor changes in the spacecraft onboard camera acquired images that is used by the CNN-based pose estimator can cause CNNs to make wrong predictions due to their reliance on low-level affected features, such as edges and textures, and their high sensitivity to slight variations in the input space. Those changes in the input images and thus on features that the CNN-based pose estimation relies on can be caused by adversarial attacks~\cite{lin2020adversarial}. Adversarial attacks aim to make small perturbations to the input images that are imperceptible to human vision and can significantly affect the CNN's prediction~\cite{grabinski2022aliasing}. For real-world applications where CNNs are applied to estimate the relative pose of spacecraft, applying an adversarial attack to the input images can potentially make the CNNs output the wrong position or attitude of the target. This could seriously damage the autonomous rendezvous operation system if wrong pose data are involved to generate any further actions, such as guidance commands for the spacecraft to rendezvous and/or dock to the target satellite.

One of the significant challenges associated with deep neural networks is that these models usually lack of transparency, which means people cannot understand how the deep neural networks achieve their decisions. To address this issue, eXplainable AI (XAI) aims to provide an understandable explanation for the AI models' decision-making process. By applying XAI methods to CNNs, such as Class Activation Mapping (CAM)~\cite{pope2019explainability}, Layer-wise Relevance Propagation (LRP)~\cite{nazari2022explainable} and SHapley Additive exPlanation (SHAP) values~\cite{lundberg2017unified}, users can understand how CNNs work and why models output their relative pose predictions. This nice characteristic of XAI methods can potentially be adopted in detecting adversarial attacks on CNN models.

This work aims to present an innovative demonstration of the vulnerability of CNN-based spacecraft rendezvous relative pose estimation scheme to digital adversarial attacks on camera input images and proposes a novel  method for detecting those adversarial attacks when they may occur. In this paper, a vision based orbital autonomous rendezvous dynamic scenario is simulated. A CNN-based pose estimator is designed and trained to estimate the relative position and attitude of the target satellite involving a modified Darknet-19~\cite{redmon2017yolo9000} as a feature extractor. The Fast Gradient Sign Method (FGSM) is employed to introduce small perturbation attacks to the input images. Various configurations of the FGSM attack are developed to demonstrate the impact of digital adversarial attacks on the CNN-based pose estimator. An LSTM-based detector exploiting the explainable Shap values of the CNN based estimator is then proposed to detect the adversarial attacks acting on the input images and thus the CNN based estimator outputs.

The paper is organised as follows: Section~\ref{related_works} provides an overview of current DL-based spacecraft pose estimation approaches and discusses existing methods for detecting adversarial attacks. Section~\ref{design} outlines the proposed design of the CNN-based pose estimator, how to adopt FGSM attacks to the pose estimator, and the design of the LSTM-based adversarial attack detector. Section~\ref{experiments} presents the test experiments that are conducted on both simulation data and real-world data obtained from our laboratory. Finally, Section~\ref{conclusion} concludes the paper and discusses future work.

\section{Background and Related Works} \label{related_works}

\subsection{DL-based Spacecraft Relative Pose Estimation}
Sharma et al.~\cite{sharma2018pose} proposed a relative pose classification network which is based on AlexNet~\cite{krizhevsky2012imagenet} architecture for non-cooperative spacecraft. In their design, the convolutional layers in AlexNet are initially trained on ImageNet dataset~\cite{deng2009imagenet} as feature extractors. The pre-trained feature extractors are adopted with two fully-connected layers and one classification layer with training on ten sets of synthetic images that were created from Tango spacecraft flown in the Prisma mission~\cite{persson2006prisma}. Their work shows that the CNN-based relative pose classification outperforms the accuracy of an architecture based on classical feature detection algorithms. However, this network is designed to output a coarse pose classification and cannot meet the precision requirements for fine position and attitude estimation missions. 

Yang et al.~\cite{yang2021pose} have proposed a CNN-based pose estimation method to estimate the relative position and orientation of non-cooperative spacecraft. In their approach, the pre-trained ResNet-50~\cite{he2016deep} is adopted as the feature extractor, and two fully-connected layers are concatenated after the feature extract to output the relative position and orientation of the target spacecraft, respectively. To adapt the network to estimate the relative pose of other similar spacecraft, an additional output layer is concatenated with the output of position and orientation to predict the category of the target spacecraft. Different from previous work introduced by Sharma et al.~\cite{sharma2018pose}, this work can output the relative position and orientation of the target spacecraft, instead of a coarse pose classification. Similarly, pre-trained ResNet has also been used as the backbone in Proen{\c{c}}a and Gao's work~\cite{proencca2020deep}. In this work, the estimation of position is achieved by two fully-connected layers with a simple regression, and the relative error is minimised based on the loss weight magnitudes. Then, the continuous attitude estimation is performed via classification with soft assignment coding~\cite{liu2011defense}. 

Rather than estimating the relative pose of spacecraft by using a single input frame, consecutive image inputs have been considered by group previous work Rondao et al., named ChiNet~\cite{rondao2022chinet}. The ChiNet featured a Recurrent Convolutional Neural Network (RCNN) architecture, which involves a modified Darknet-19~\cite{redmon2017yolo9000} as an image feature extractor and followed by LSTM units to deal with the sequences of input images. The ChiNet takes 4-channels input which not only includes the RGB image but also a thermal image of the spacecraft that has been stacked to the fourth channel of input. The ChiNet also proposed a multistage optimisation approach to train the deep neural network to improve the performance in spacecraft relative pose estimation.

\subsection{Explainability in CNNs}
While recent approaches to DL-based spacecraft relative pose estimation demonstrate outstanding performance in terms of prediction accuracy, understanding how these models predict relative pose is essential for providing robust solutions for future space rendezvous missions. As a new approach  solution, eXplainable AI (XAI) techniques offer the possibility to analyse gradients in DL models to indicate the significance of input variables in the estimation decision-making process.

Lundberg and Lee proposed the SHAP values to interpret complex machine learning models~\cite{lundberg2017unified}.  The SHAP value is based on a concept from game theory called Shapley values. These are used to fairly distribute the payoff among the players of a cooperative game, where each player can have different skills and contributions. Similarly, SHAP values assign each feature an importance value for a particular prediction and provide insights into the contribution of each feature. By examining the SHAP values of machine learning models, we will able to understand the predictions of complex machine learning models.

Contrastive gradient-based (CG) saliency maps~\cite{simonyan2013deep} are visual explanation methods for deep neural networks. They produce a heat map where the norm of the model's gradients indicates the significance of input variables. The heat map highlights the areas in the input image that would change the output class if they were changed. By accessing the heat map, users can identify the most relevant features for the model’s prediction. 

Class Activation Mapping (CAM)~\cite{zhou2016learning} generates visual explanation maps by finding the spatial locations in the input image that contribute the most to a specific prediction. The CAM is particularly helpful in image classification tasks through CNNs. Similarly, gradient-weighted CAM (Grad-CAM)~\cite{selvaraju2017grad} extends the work of CAM and provides visual explanations for decisions made by a wide range of CNN-based methods. Grad-CAM utilises the gradients of any target concept, flowing into the final convolutional layer to produce a localisation map that highlights the important regions in the input image for predicting the concept.  These XAI methods interpret the CNNs, making people understand how and why CNNs make certain predictions. However, since then, there has been no specific analysis on interpreting the DL-based spacecraft relative pose estimation to improve their explainability.

\subsection{Adversarial Attacks}
Adversarial attacks for CNNs aim to make small perturbations on the original input images where original and perturbed images look similar in human vision but can significantly impact the CNNs' predictions. However, very limited research works are investigating how adversarial attacks can impact DL-based pose estimation systems. Chawla et al.~\cite{chawla2022adversarial} demonstrate the effect of different types of adversarial attacks on the predictions of the DL-based pose estimation system. Their work shows that adversarial attacks can significantly impact monocular pose estimation networks, leading to increased trajectory drift and altered geometry. Similarly, Nemcovsky et al~\cite{nemcovsky2022physical} illustrate that the physical passive path adversarial attacks can seriously increase the error margin of a visual odometry model which is used in autonomous navigation systems leading onto potential collisions. 

To protect the DL-based system from adversarial attacks, Liu et al~\cite{liu2020towards} proposed a detection method based on the robustness of the classification results. Their results show that the detector performs well against gradient-based adversarial attacks. Our group work, Hickling et al~\cite{hickling2023robust}, proposed a CNN-based adversarial attack detector and an LSTM-based adversarial attack detector for Deep Reinforcement Learning (DRL) based Uncrewed Aerial Vehicle guidance. The simulation results show that the LSTM-based adversarial attack detector leads to 90\% detection accuracy on the DRL model. It also suggests that the LSTM-based detector performs much more accurately and quicker than the CNN-based adversarial attack detector. Indeed, the LSTM-based detector is demonstrated to meet the real-time requirement in DRL based guidance.

To the best of our knowledge, as of yet, there is no literature looking at the impact of adversarial attacks in spacecraft relative pose estimation and how to detect those adversarial attacks in DL-based spacecraft relative pose estimation systems and this work first time proposes this. Our objective is to ultimately create an adversarial attack detector for the space navigation system, which employs SHAP values explainability mechanism to detect and flag potential adversarial attacks.

\section{Methodology} \label{design}

In this section, a CNN-based spacecraft relative pose estimator is newly designed with the aim of providing a reliable estimated position and attitude of the target spacecraft in as rendezvous scenario. Then, the FGSM attacks are adopted on the spacecraft onboard camera resulting in an adversarial image to evaluate the impacts on the proposed deep pose estimator. Next, SHAP values are introduced to generate XAI signatures for both adversarial and normal input images. Finally, an LSTM-based adversarial detector is proposed and trained, which learns normal and adversarial SHAP values to detect the adversarial attacks on the spacecraft relative pose estimator.

\subsection{CNN-based Spacecraft Relative Pose estimator}
\subsubsection{Overall architecture design}
Similarly, as most DL-based spacecraft relative pose estimation algorithms, CNN is proposed for extracting features in the proposed pose estimator. The overall design of the pose estimator follows the design methodology in ChiNet~\cite{rondao2022chinet}. The Darknet-19~\cite{redmon2017yolo9000} is introduced as the feature extractor in this design. The Darknet-19 is originally trained in ImageNet~\cite{deng2009imagenet} dataset which has an input size of $244 \times 244$. In our design, input images of the pose estimator have a larger size than ImageNet images. Therefore, the first convolutional layer in Darknet-19 is configured with a kernel size of $7 \times 7$. Following the approach of Darknet-53~\cite{redmon2018yolov3}, the maxpooling layers in the Darknet-19 are replaced by $3 \times 3$ convolution operation with a stride of 2. Similarly, as the Darknet-53 approaches, the residual connections are also adopted to the proposed pose estimator. Batch Normalisation~\cite{ioffe2015batch} layers are applied after each convolutional layer.

\begin{figure*}[htbp!]
    \centering
    \includegraphics[scale=0.4]{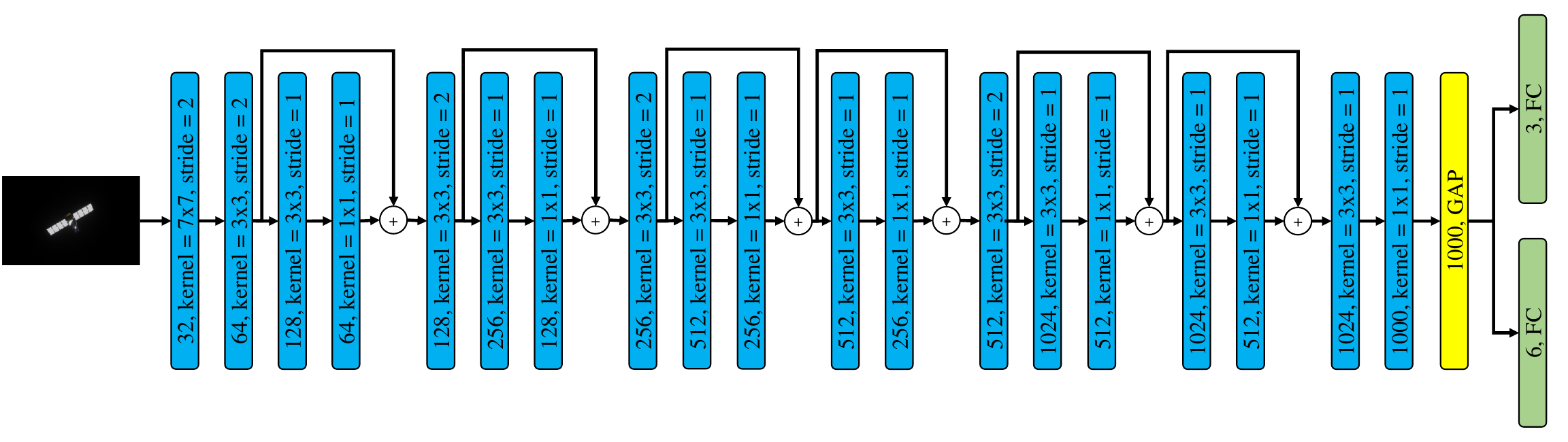}
    \caption{The overall architecture of the proposed spacecraft relative pose estimator. The blue blocks represent the convolutional layers, which are formatted as $(layer~size, kernel~size, stride)$). Each convolutional layer is followed by a batch normalization layer and LeakyReLu activation. The yellow block indicates the Global Average Pooling (GAP) layer that downsamples the exacted features to a fixed 1D vector of 1000 units. The green blocks represent FC layers that will output the estimated relative position and attitude, respectively. }
    \label{fig:pose estimator}
\end{figure*}

Our deep spacecraft relative pose estimator aims to output the relative position and attitude of the target directly. Therefore, two separate FC layers are applied. The first FC layer involves 3 output nodes to output the relative position in \((x, y, z)\) and the second FC layer adopts a 6-dimensional (6-D) vector to represent the relative attitude of the target spacecraft. Finally, two FC layers are concatenated together to output the relative 6-DOF pose. In the second FC layer, 6-D vectors are applied to represent the relative attitude of the target spacecraft, instead of using quaternion representation. The reason is that the relative pose estimator is designed as a regression problem where the output has to be continuous. However, the normal attitude representation of quaternion is discontinuous, due to its antipodal ambiguity, i.e. $-q = q$. Therefore, the proposed pose estimator applies the 6-D vector formulated by Zhou et al.~\cite{zhou2019continuity}, which mapped the 3-dimensional rotations into a 6-D continuous rotation. The overall design of the spacecraft relative pose estimator is presented in Fig.~\ref{fig:pose estimator}.

\subsubsection{Synthetic data generation}
To train and test the spacecraft relative pose estimator, synthetic datasets are generated in Blender, which is an open-source 3D modelling software. The spacecraft target model used in the synthetic dataset generation is the Jason-1 satellite, which was downloaded from the NASA 3D Resources website~\cite{NASA_Jason1}. Dynamic simulation of the rendezvous is developed to generate the synthetic dataset in which the camera onboard the chaser spacecraft starts at 60 metres away in \(z-axis\) from the target and end at 10 metres away from the target in \(z-axis\), i.e. (0,0,10). Random rotation of the camera and target is considered in the synthetic data generation. Many trajectory sequences are generated and each sequence contains 2,500 images. To prevent overfitting in the deep relative pose estimator network, random rotation of the target spacecraft is applied to the model, and the camera is initialised at various positions in the synthetic data generation. Table~\ref{tab:synthetic data gene} illustrates the synthetic dataset generated for training and validating the deep pose estimator. 

\begin{table}[htbp!]
    
    \caption{Example of synthetic data generated from Blender}
    \begin{center}
    \begin{tabular}{c c c }
        \hline
           Sequence ID  &   Start Position  &   Target Rotation \\
       \hline
               0        &   (0,0,60)        &       0           \\
               1        &   (-15,-25,60)    &       0           \\
               2        &   (-15,25,60)     &       0           \\
               3        &   (15,25,60)      &       0           \\
               4        &   (15,-25,60)     &       0           \\
               5        &   (-15,-10,60)    &   $\pm$ 10 deg    \\
               6        &   (-15,10,60)     &   $\pm$ 10 deg    \\
               7        &   (15,10,60)      &   $\pm$ 10 deg    \\
               8        &   (15,-10,60)     &   $\pm$ 10 deg    \\
               9        &   (-15,-10,60)    &   $\pm$ 10 deg    \\
               10       &   (-15,10,60)     &   $\pm$ 10 deg    \\
               11       &   (15,10,60)      &   $\pm$ 10 deg    \\
               12       &   (15,-10,60)     &   $\pm$ 10 deg    \\
        \hline
    \end{tabular}
    \end{center}
    \label{tab:synthetic data gene}
\end{table}

\subsubsection{Loss Function}
Training the spacecraft relative pose estimator can be formulated as a regression problem, where the total loss function combines the loss in position and loss in attitude. These are computed by \eqref{position_loss} and \eqref{attitude_loss}, respectively, which were originally proposed by Kendall et al~\cite{kendall2018multi}. Followed by Rondao et al~\cite{rondao2022chinet}, a trainable weight is attributed to each loss, which corresponds to the task-specific variance of the Gaussian distribution. The total loss is then formulated in \eqref{total_loss}.

\begin{equation}\label{position_loss}
    L_{p} = \sum_{i = 0}^{B}(|| p_{pred}^{i}-p_{gt}^{i} ||)
\end{equation}

\begin{equation}\label{attitude_loss}
    L_{r} = \sum_{i = 0}^{B}(|| r_{pred}^{i}-r_{gt}^{i} ||)
\end{equation}

\begin{equation}\label{total_loss}
    L_{total} = exp(-2\sigma_{p}) L_{p} + exp(-2\sigma_{r})L_{r} + 2(\sigma_{p}+\sigma_{r})
\end{equation}
where the $p_{pred}$ and $r_{pred}$ indicate the predicted position and attitude, and $p_{gt}$ and $r_{gt}$ indicate the ground truths position and attitude, respectively. $B$ is the batch size and $||\cdot||$ donates the $L_{2}$ norm. $\sigma_{p}$ and $\sigma_{r}$ represent the learnable weights for position and attitude, respectively.

\subsection{Adversarial Attacks}
In this work, the adversarial examples are generated by FGSM attacks~\cite{goodfellow2014explaining}. The FGSM attacks aim to add small perturbations to the input images which will maximise the network's loss. The equation in \eqref{fgsm} describes how to generate an adversarial example for a given input image $x$ by FGSM attack.
\begin{equation}\label{fgsm}
    x' = x + \epsilon \times sign(\nabla_{x}L(\theta,x,y))
\end{equation}
where $\epsilon$ is a value of the perturbation effect which describes how strong the attack is. $L$ is the loss of the input $x$ with ground truth of $y$ . The $(\nabla_{x}$ calculates the loss gradient, $L$ for input image $x$ with relative ground truth $y$, and $\theta$ indicates the trained network's parameters. Depending on the quality of input images and the attack strength, the result of the FGSM attack can be modified by changing the $\epsilon$ value. 

\begin{figure*}
  \centering

  \subfigure[]{%
    \includegraphics[width=0.28\textwidth]{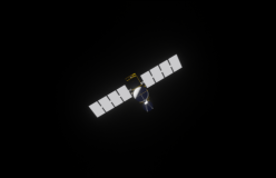}
    \label{fig:originalExample}
  }
  \raisebox{0.08\linewidth}{\includegraphics[width=0.03\textwidth]{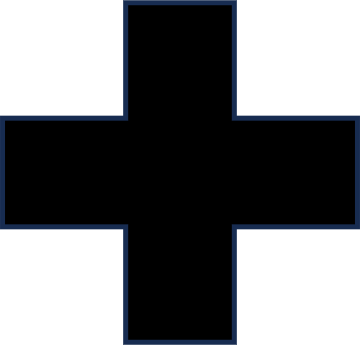}}
  \subfigure[]{%
    \includegraphics[width=0.28\textwidth]{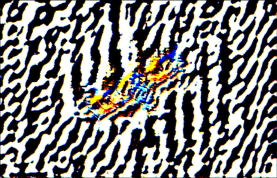}
    \label{fig:perturbationPatch}
  }
  \raisebox{0.08\linewidth}{\includegraphics[width=0.03\textwidth]{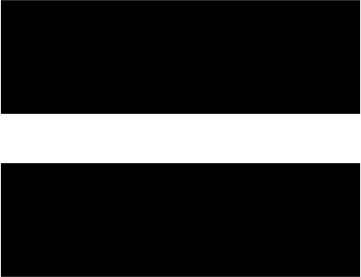}}
  \subfigure[]{%
    \includegraphics[width=0.28\textwidth]{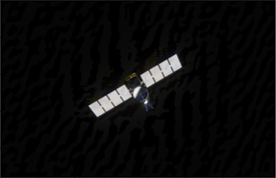}
    \label{fig:advImg}
  }

  \caption{An example of applying FGSM attacks to the input image. (a) the original input image. (b) perturbation patch with $\epsilon=0.05$. (c) resultant adversarial image.}
  \label{fig: adversarial example}
\end{figure*}

In real implementation, the $\epsilon$ needs to be small enough to ensure the perturbations on the input image are seamless and cannot be visible by human vision but still significantly change the deep model's predictions. The $\epsilon$ value should be in the range of (0,1), where a value of 0 means the adversarial image will be the same as the input image without any perturbation and a value of 1 means the adversarial image will be perturbed as significant distorted image to human vision. Fig.~\ref{fig: adversarial example} illustrates an example of applying FGSM attacks to input images of the spacecraft relative deep pose estimator. 

\subsection{Explanability and Adversarial Attacks Detector}
\subsubsection{Explanability via DeepSHAP}
The black-box nature of deep neural networks makes users can only observe the prediction of these models, but do not know the reasons for getting correct or wrong predictions. XAI techniques are developed to interpret the DL models. When the model's prediction is changed, the XAI will generate relative explanations to explain why the model is getting the prediction.  In this work, we proposed a novel approach that adopts XAI techniques by applying the change in SHAP values of the input images as a measure to determine whether an adversarial attack happens on input images.

Originally, SHAP is proposed based on the idea of Shapley values, which are designed to assign a credit to every input feature for a given prediction. Generating SHAP values for DNNs can be computationally expensive, as the DNNs normally contain a huge amount of features. 
Thanks to the work of DeepLIFT~\cite{shrikumar2017learning}, Shapley values for DNNs can be estimated by linearising the non-linear components of a neural network, a method referred to as DeepSHAP~\cite{lundberg2017unified}. This is achieved by utilising a reference input distribution, which can be linearly approximated, to estimate the expected value for the entire model. 

However, directly generating SHAP values for the spacecraft relative pose estimator still requires a large amount of computational resources. The pose estimator is based on CNNs with image inputs that contain thousands of pixels. Using DeepSHAP for image input requires generating Shapley values for each single pixel for every output neuron. Therefore, in this work, we consider computing the SHAP values for the subsampling layer in the pose estimator, instead of computing them for the input image. As demonstrated previously, the spacecraft relative pose estimator contains a GAP layer that downsamples feature maps from the prior convolutional layer to 1000 samples. For example, computing SHAP values for a $720 \times 480$ image needs to compute 355,200 pixels, instead, the GAP layer in the pose estimator only employs 1000 neurons. As a result, SHAP values are generated for the outputs of the GAP layer that only need to compute 1000 features. This saving in the computation makes the generation of SHAP values for the deep pose estimator could potentially meet the implementation time constraints.

\subsubsection{Adversarial attack detector}

To detect any incoming adversarial attacks on the spacecraft deep relative pose estimator trhough the onboard camera, an LSTM-based adversarial attacks detector is proposed. The detector aims to monitor the SHAP values generated from the output of the GAP layer and detect any slight anomaly changes that could result based an adversarial attack. The LSTM is a type of Recurrent Neural Networks (RNNs) that is widely used in learning from time-series data, such as speech recognition~\cite{yu2019review}. The LSTM architecture was originally proposed to address the long-term dependency issue in conventional RNNs. It can enable the propagation and representation of information over a sequence without causing useful information from distant past time steps to be ignored. 

In our approach, the SHAP values are generated based on the prediction of each output neuron in the proposed deep pose estimator. Different from applying adversarial attacks on a classification CNN that only change the output label, when an attack occurs on the deep pose estimator, it could affect all output neurons to estimate for wrong position and attitude. Therefore, it can be assumed that there might exist a certain level of dependencies among those output neurons. From this point of view, building an LSTM-based adversarial attack detector can potentially achieve high detection accuracy.

\begin{figure}[htbp!]
    \centering
    \includegraphics[width=\linewidth]{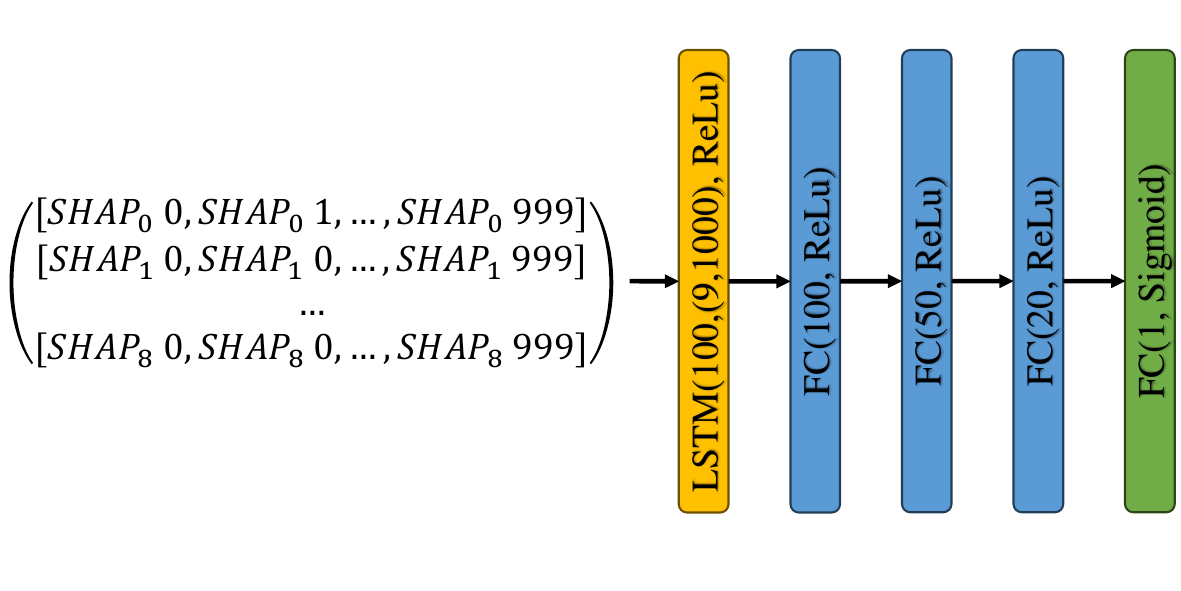}
    \caption{Proposed adversarial attack detector. The yellow block indicates the LSTM layer which has an input shape of (9,1000) and an output space of 100. ReLu is applied as the activation function for the LSTM layer. The blue blocks are FC layers in the format of $(units,~activation)$. The green block indicates the output layer of the adversarial detector, which is also formed from the FC layer and outputs a Boolean to detect adversarial attacks.}
    \label{fig:detector}
\end{figure}

Fig.~\ref{fig:detector} introduces the architecture of the proposed adversarial attack detector. The detector takes the SHAP values that are computed from the GAP layer of the deep pose estimator. As there are nine output neurons in the proposed deep pose estimator, the shape of the SHAP values is $(9,1000)$. To input SHAP values to the detector, the SHAP values are formatted as a sequence data with a length of 9. The detector outputs a Boolean, $True/False$, to indicate the result of detecting adversarial attacks.

\section{Experimental Results} \label{experiments}

To validate our adversarial detection approach, two experiments are performed. The first experiment is built on the simulation environment with synthetic data as mentioned in Section~\ref{design}. The second experiment is built on our lab environment to  testing our approach with real data. For both sets of experiments, the spacecraft deep relative pose estimator and the adversarial attack detector are tested for their relevant accuracy, and then the two systems are integrated to test the overall successful detection rate of adversarial attacks.

\subsection{Results on Synthetic Data}
\subsubsection{Accuracy of the Spacecraft Deep Relative Pose Estimator}
\begin{figure}[h!]
  \centering

  \subfigure[]{%
    \includegraphics[width=0.48\linewidth]{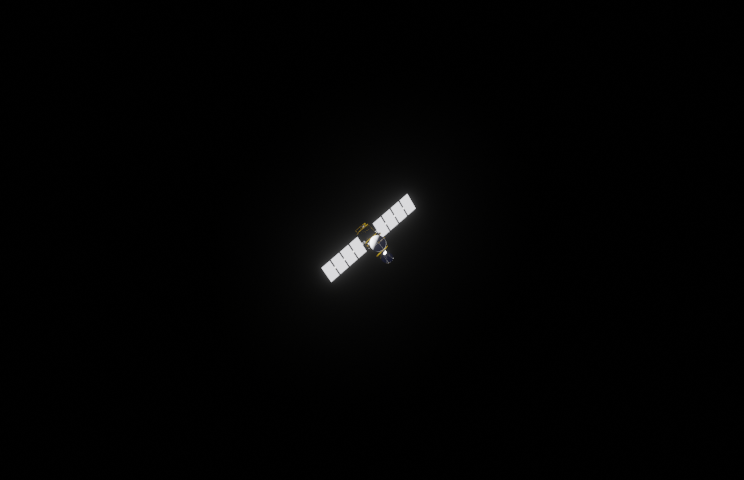}
    \label{fig:60m}
  }\hfill
  \subfigure[]{%
    \includegraphics[width=0.48\linewidth]{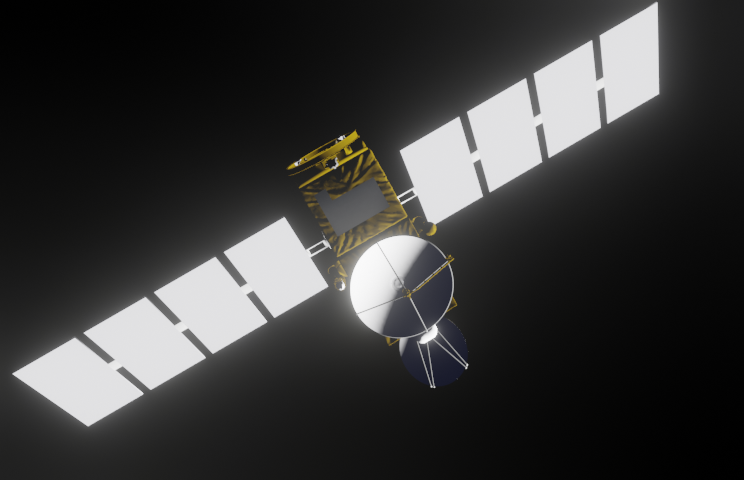}
    \label{fig:10m}
  }

  \caption{Examples of synthetic data generated from Blender. (a) image captured at a camera position of (0, 0, 60m). (b) image captured at a camera position of (0, 0, 10m). Random rotation is applied to the target spacecraft.}
  \label{fig:synthetic example.}
\end{figure}

To train the deep relative pose estimator, image data are collected from the Blender 3D model. There are 13 sequences of images generated from Blender with the relevant trajectories that are mentioned in Table~\ref{tab:synthetic data gene}. By following the trajectories in Table~\ref{tab:synthetic data gene}, 2,500 images are generated for each trajectory, resulting in a dataset with 32,500 images for training and testing in total. Fig.~\ref{fig:synthetic example.} shows two examples of synthetic data generation in Blender. 

The synthetic dataset is split by a train-test ratio of 0.8, i.e. 80\% of images in the dataset are used for training the deep relative pose estimator, and 20\% of images are used to test the model's accuracy. Each image is associated with a ground truth label in the format of $(x,y,z,w,x_{i},y_{j},z_{k})$. The first three elements in the ground truth label represent the relative position of the chaser onboard camera to the target and the rest 4 elements represent the target attitude in quaternion representations in the chaser camera frame. The deep relative pose estimator outputs the attitude in a 6-D vector. Therefore, to calculate the loss in attitude, the quaternion representations are converted to the 6-D vector representation by following the approach in~\cite{zhou2019continuity}. A dropout rate of 0.2 is applied to the GAP layer in the training process. Multiple data augmentation techniques are considered in training the deep relative pose estimator, including Gaussian blur, Gaussian noise, image compression, random brightness and so on. These techniques help to prevent the model from overfitting the training dataset. The deep relative pose estimator is trained by stochastic gradient descent with an Adam optimiser. The triangular2~\cite{smith2017cyclical} policy is applied for cycling learning rate with base learning of 2.5e-5. 

\begin{figure}[h!]
    \centering
    \includegraphics[width=\linewidth]{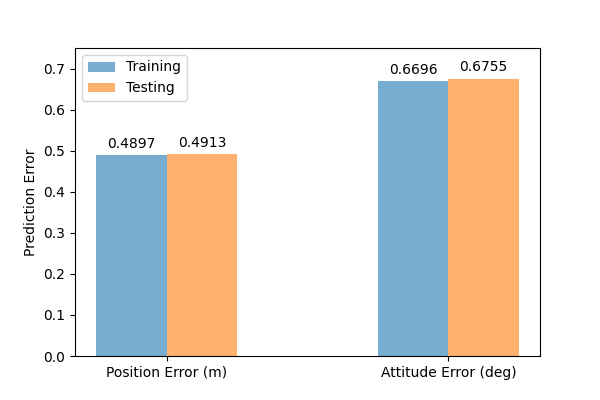}
    \caption{The prediction accuracy of the proposed pose estimator on training and test dataset after 50 epochs. The blue bar presents the average error on training data and the orange bar represents the average error on test data}
    \label{fig:syn train results}
\end{figure}

After training the deep relative pose estimator for 50 epochs with the training batch size of 32, The model's average prediction accuracy for both training and test datasets is reported in Fig.~\ref{fig:syn train results}. In this experiment, the position error is measured by~\eqref{eq: pos err} and the attitude error is measured by~\eqref{eq: rot err}.
\begin{equation}\label{eq: pos err}
    p_{err} := ||p_{pred} - p_{gt}||
\end{equation}
\begin{equation}\label{eq: rot err}
    r_{err} := 2\arccos(q^{-1}_{pred}\otimes q_{gt})
\end{equation}
where $p_{pred}$ and $p_{gt}$ represent the prediction of position and the ground truth of position magnitude. The $q_{pred}$ and $q_{gt}$ indicate the prediction of attitude and the ground truth of attitude in quaternion representation. The $\otimes$ denotes the quaternion multiplication and $||\cdot||$ denotes the $L_{2}$ norm.

The proposed spacecraft relative pose estimator achieves an accuracy of around 0.49m in position error and 0.68 deg in attitude error on the test dataset. Table~\ref{tab:compare} reports a comparison between the proposed deep relative pose estimator and state-of-the-art performance of other DL-based space relative pose estimation approaches based on their datasets. The comparison here aims to show that the proposed spacecraft deep relative pose estimator can achieve relatively good performance on the synthetic data and can be used as a baseline model to implement the adversarial attack algorithm on and test the adversarial attack detector. The comparison is not meant to be a quantitative benchmark evaluation of our approach relative to existing performing approaches.
\begin{table}[h!]
    \centering
    
    \caption{Comparison with other approaches in DL-based space relative pose estimation }
    \resizebox{\linewidth}{!}{\begin{tabular}{cccc}
        \hline
           Model  &   Dataset  &   Position Error (m) &   Attitude Error (deg)      \\
        \hline
           \cite{proencca2020deep}  &   SPEED~\cite{kelvins}   &   0.56   &   8.0   \\
           \cite{rondao2022chinet}  &   Synthetic   &   1.73   &   6.62             \\  
           \cite{yang2021pose}      &   Synthetic   &   [0.052, 0.039, 0.077] &  [0.213, 0.233, 0.097]\\
           Ours                     &   Synthetic   &   0.49    &   0.68            \\
        \hline
    \end{tabular}}
    \label{tab:compare}
\end{table}

\subsubsection{FGSM Adversarial Attacks}
\begin{figure}[h!]
    \centering
    \includegraphics[width=\linewidth]{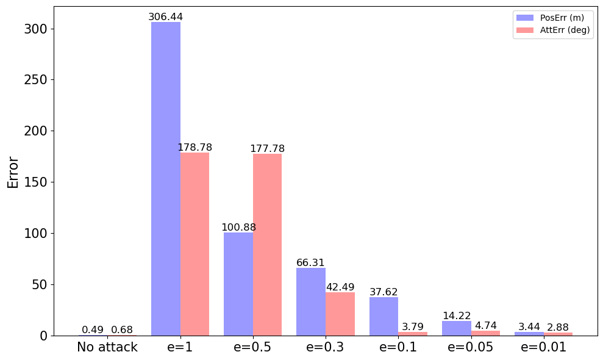}
    \caption{Comparison of the prediction error of pose estimator under FGSM attack on test data with various $\epsilon$ values. The blue bar indicates the average position error and the red bar indicates the average attitude error on test data. The error magnitude for the position is metres and the error magnitude in rotation is measured by degrees.}
    \label{fig:fgsm test results}
\end{figure}

As discussed in Section~\ref{design}, the perturbation made by FGSM attacks can be adjusted by changing the $\epsilon$ value. To investigate the impact of adversarial attacks on DL-based space relative pose estimation, different $\epsilon$ values are selected to generate adversarial onboard camera image input to the proposed deep relative pose estimator. Typically, the $\epsilon$ applied in this experiment are 1, 0.5, 0.3, 0.1, 0.05 and 0.01. The larger value of $\epsilon$ is, the more perturbations are made to images. The FSGM attack is applied to all synthetic test data, where all images in the test data. Then, the perturbed images are fed to the deep relative pose estimator for testing the impact of the FGSM attack. The average prediction relative pose errors of applying different $\epsilon$ values are reported in Fig.~\ref{fig:fgsm test results}.

We can see that as the $\epsilon$ value increases, the deep model's prediction error becomes larger. The attitude error is quite stable on $\epsilon = 0.1,0.05$ and $0.01$, but has a dramatic increase if the $\epsilon > 0.3$. 

\begin{figure}[h!]
    \centering
    \includegraphics[width=\linewidth]{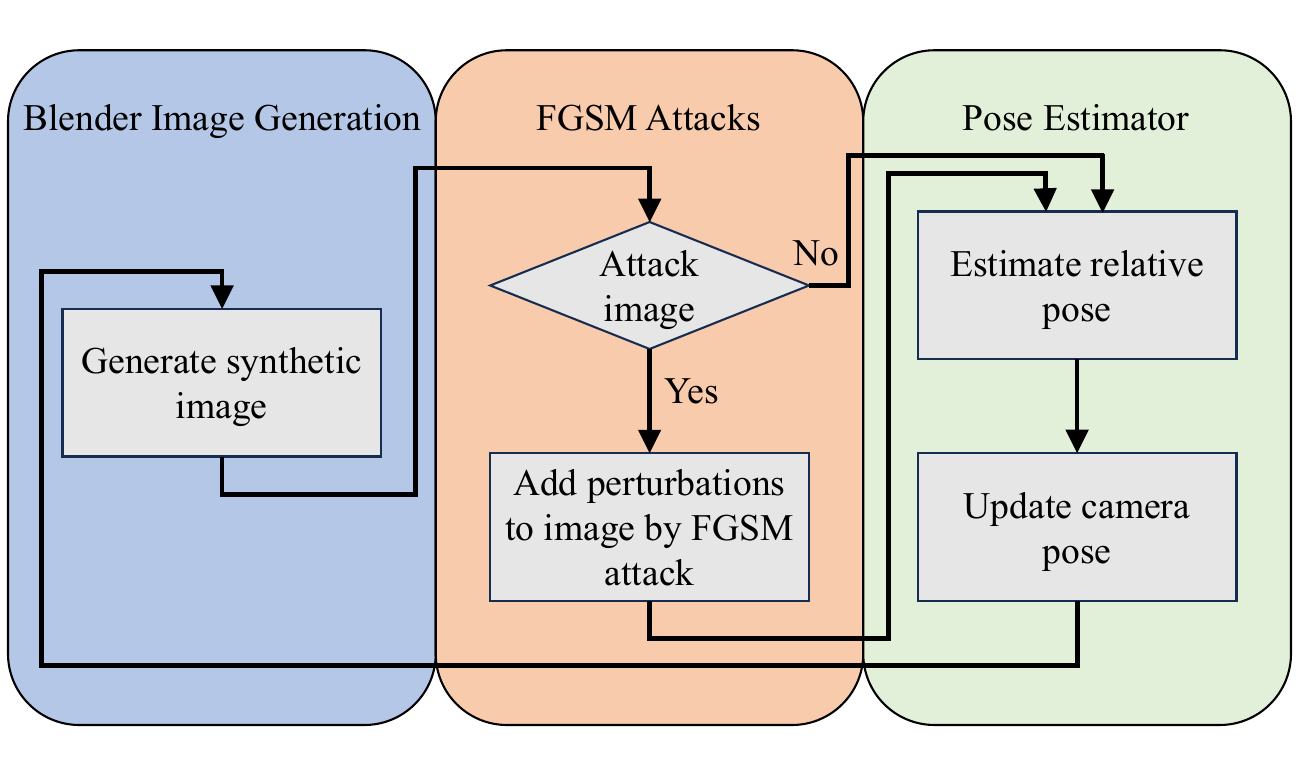}
    \caption{Test system for proposed pose estimator on Blender in simulated space rendezvous scenario.}
    \label{fig:test loop}
\end{figure}

To assess well how the adversarial attack can impact the DL-based navigation system in a space rendezvous scenario, a simple guidance scheme is implemented with the proposed deep relative pose estimator. The guidance scheme takes the estimated relative pose from the proposed deep relative pose estimator and then provides relative control actions to move the camera (spacecraft) to the target position. In the guidance scheme, the camera has an initial position of $(0,0,60)$ and a target position of $(0,0,10)$ with $\pm 1m$ tolerance. The guidance scheme updates the camera position with a maximum velocity of 1$m/s$, as described in~\eqref{eq:update} and~\eqref{eq:difference}
\begin{equation}\label{eq:update}
    p_{new} = 
    \begin{cases}
        p_{est} -1 & \text{if } diff \geq 1 \\    
        p_{est} -diff & \text{otherwise}
    \end{cases}
\end{equation}
\begin{equation}\label{eq:difference}
    diff = p_{est} - p_{tar}
\end{equation}
where $p_{new}$, $p_{est}$, $p_{tar}$ present the updated position, estimated position and target position of the camera, respectively. The test system is implemented as shown in Fig.~\ref{fig:test loop}. 

In this experiment, the test system is continuously attacked by FGSM on image data with various acquired camera frames. The $success$ attack is defined as the camera (spacecraft) missing the target position while the $failure$ attack means that the camera (spacecraft) can still reach the target position under continuous FGSM attack. Experimental results are reported in Table~\ref{tab:e=0.5} - \ref{tab:e=0.01}. 
\begin{table}[htbp!]
\centering
\caption{FGSM attacks on the simulated space rendezvous scenario with $\epsilon$=0.5}
\resizebox{\linewidth}{!}{\begin{tabular}{cccccc}
\hline
\multicolumn{6}{c}{$\epsilon$ = 0.5} \\ 
\hline
\multicolumn{1}{c|}{\begin{tabular}[c]{@{}c@{}}Continuously \\Attacked  Frame\end{tabular}} 
&  & 5          & 10         & 15         & 20         \\ 
\hline
\multicolumn{1}{c|}{\multirow{6}{*}{\begin{tabular}[c]{@{}c@{}}Attack start point\\    \\ (m)\end{tabular}}} 
& 60 & failure    & failure    & failure    & Success \\
\multicolumn{1}{c|}{}                                                                                        
& 50 & failure    & failure    & failure    & Success \\
\multicolumn{1}{c|}{}                                                                                        
& 40 & failure    & failure    & Success    & Success \\
\multicolumn{1}{c|}{}                                                                                        
& 30 & failure    & failure    & Success    & Success \\
\multicolumn{1}{c|}{}                                                                                        
& 20 & failure    & Success    & Success    & Success \\
\multicolumn{1}{c|}{}                                                                                        
& 10 & Success    & Success    & Success    & Success \\ 
\hline
\end{tabular}}
\label{tab:e=0.5}
\end{table}

\begin{table}[htbp!]
\centering
\caption{FGSM attacks on the simulated space rendezvous scenario with $\epsilon$=0.3}
\resizebox{\linewidth}{!}{\begin{tabular}{cccccc}
\hline
\multicolumn{6}{c}{$\epsilon$ = 0.3} \\ 
\hline
\multicolumn{1}{c|}{\begin{tabular}[c]{@{}c@{}}Continuously \\Attacked  Frame\end{tabular}} 
&  & 5          & 10         & 15         & 20         \\ 
\hline
\multicolumn{1}{c|}{\multirow{6}{*}{\begin{tabular}[c]{@{}c@{}}Attack start point\\    \\ (m)\end{tabular}}} 
& 60 &          & failure    & failure    & Success \\
\multicolumn{1}{c|}{}                                                                                        
& 50 &          & failure    & Success    & Success \\
\multicolumn{1}{c|}{}                                                                                        
& 40 &          & failure    & Success    & Success \\
\multicolumn{1}{c|}{}                                                                                        
& 30 &          & failure    & Success    & Success \\
\multicolumn{1}{c|}{}                                                                                        
& 20 & failure  & Success    & Success    & Success \\
\multicolumn{1}{c|}{}                                                                                        
& 10 & failure  & Success    & Success    & Success \\ 
\hline
\end{tabular}}
\label{tab:e=0.3}
\end{table}

\begin{table}[htbp!]
\centering
\caption{FGSM attacks on the simulated space rendezvous scenario with $\epsilon$=0.1}
\resizebox{\linewidth}{!}{\begin{tabular}{cccccc}
\hline
\multicolumn{6}{c}{$\epsilon$ = 0.1} \\ 
\hline
\multicolumn{1}{c|}{\begin{tabular}[c]{@{}c@{}}Continuously \\Attacked  Frame\end{tabular}} 
&  & 5          & 10         & 15         & 20         \\ 
\hline
\multicolumn{1}{c|}{\multirow{6}{*}{\begin{tabular}[c]{@{}c@{}}Attack start point\\    \\ (m)\end{tabular}}} 
& 60 &          &            &            & failure \\
\multicolumn{1}{c|}{}                                                                                        
& 50 &          &            &            & failure \\
\multicolumn{1}{c|}{}                                                                                        
& 40 &          &            & failure    & Success \\
\multicolumn{1}{c|}{}                                                                                        
& 30 &          &            & failure    & failure \\
\multicolumn{1}{c|}{}                                                                                        
& 20 &          &            & failure    & failure \\
\multicolumn{1}{c|}{}                                                                                        
& 10 &          &            & failure    & Success \\ 
\hline
\end{tabular}}
\label{tab:e=0.1}
\end{table}

\begin{table}[htbp!]
\centering
\caption{FGSM attacks on the simulated space rendezvous scenario with $\epsilon$=0.05}
\resizebox{\linewidth}{!}{\begin{tabular}{cccccc}
\hline
\multicolumn{6}{c}{$\epsilon$ = 0.05} \\ 
\hline
\multicolumn{1}{c|}{\begin{tabular}[c]{@{}c@{}}Continuously \\Attacked  Frame\end{tabular}} 
&  & 5          & 10         & 15         & 20         \\ 
\hline
\multicolumn{1}{c|}{\multirow{6}{*}{\begin{tabular}[c]{@{}c@{}}Attack start point\\    \\ (m)\end{tabular}}} 
& 60 &          &            & failure    & failure \\
\multicolumn{1}{c|}{}                                                                                        
& 50 &          &            & failure    & Success \\
\multicolumn{1}{c|}{}                                                                                        
& 40 &          & failure    & failure    & Success \\
\multicolumn{1}{c|}{}                                                                                        
& 30 &          & failure    & Success    & Success \\
\multicolumn{1}{c|}{}                                                                                        
& 20 & failure  & Success    & Success    & Success \\
\multicolumn{1}{c|}{}                                                                                        
& 10 & failure  & Success    & Success    & Success \\ 
\hline
\end{tabular}}
\label{tab:e=0.05}
\end{table}

\begin{table}[htbp!]
\centering
\caption{FGSM attacks on the simulated space rendezvous scenario with $\epsilon$=0.01}
\resizebox{\linewidth}{!}{\begin{tabular}{cccccc}
\hline
\multicolumn{6}{c}{$\epsilon$ = 0.01} \\ 
\hline
\multicolumn{1}{c|}{\begin{tabular}[c]{@{}c@{}}Continuously \\Attacked  Frame\end{tabular}} 
&  & 5          & 10         & 15         & 20         \\ 
\hline
\multicolumn{1}{c|}{\multirow{6}{*}{\begin{tabular}[c]{@{}c@{}}Attack start point\\    \\ (m)\end{tabular}}} 
& 60 &          &            & failure    & failure \\
\multicolumn{1}{c|}{}                                                                                        
& 50 &          &            & failure    & failure \\
\multicolumn{1}{c|}{}                                                                                        
& 40 &          &            & failure    & failure \\
\multicolumn{1}{c|}{}                                                                                        
& 30 &          &            & failure    & failure \\
\multicolumn{1}{c|}{}                                                                                        
& 20 &          &            & failure    & failure \\
\multicolumn{1}{c|}{}                                                                                        
& 10 &          &            & failure    & failure \\ 
\hline
\end{tabular}}
\label{tab:e=0.01}
\end{table}

From Table~\ref{tab:e=0.5} - \ref{tab:e=0.01}, we can clearly see that the adversarial attack can result in a significant impact on the guidance based DNN-based relative navigator typically when the distance between the camera and the target is smaller than 30m. In most cases, continuously attacking the deep model for more than 15 frames after the camera approaches less than 30m to the target, the camera (spacecraft) will fail to reach the target position. In a real space rendezvous mission where a chaser relies on a DL-based relative pose estimation system, an adversarial attack has the potential to cause the chaser to fail in approaching the target position, resulting in mission failure. Therefore, detecting adversarial attacks on DL-based pose estimators becomes critical.

\subsubsection{LSTM-based Adversarial Attack Detector}

The proposed adversarial attack detector is designed based on the LSTM architecture. It aims to detect the change in SHAP values when an adversarial attack occurs on the input image. As mentioned in Section~\ref{design}, the SHAP values are computed at the output of the GAP layer in the proposed deep relative pose estimator. The GAP layer contains 1000 neurons, therefore, 1000 values are calculated for each output neuron, resulting 9$\times$1000 output SHAP values.  

In our approach, the SHAP values of the GAP layer are calculated by DeepSHAP~\cite{lundberg2017unified} algorithm. The DeepSHAP algorithm computes SHAP values for inputs by integrating over background samples. It then estimates approximate SHAP values in a manner that sums up the difference between the expected deep model's output on the background samples and the current model's output. In this work, 1000 images are randomly selected from the training dataset to compute the downsampled features at the GAP layer. These samples serve as the background samples for the DeepSHAP explainer. To train the adversarial attack detector, we generated 15,000 sets of SHAP values for normal samples and an additional 15,000 sets of SHAP values for adversarial samples. The normal samples consist of the entire test dataset, which is used for testing the deep pose estimator, along with a random selection of images from the training dataset. This random selection was made to reach a total of 15,000 samples, thereby bridging the gap between this number and the number of images in the test dataset by the deep relative pose estimator. Adversarial samples are generated by attacking the test dataset of the deep relative pose estimator with various $\epsilon$ values: 0.5, 0.3, 0.1, 0.05, and 0.01. Subsequently, 3,000 perturbed images are randomly selected from each $\epsilon$ value for calculating the corresponding SHAP values.

\begin{figure*}[htbp!]
    \centering
    \includegraphics[width=\linewidth]{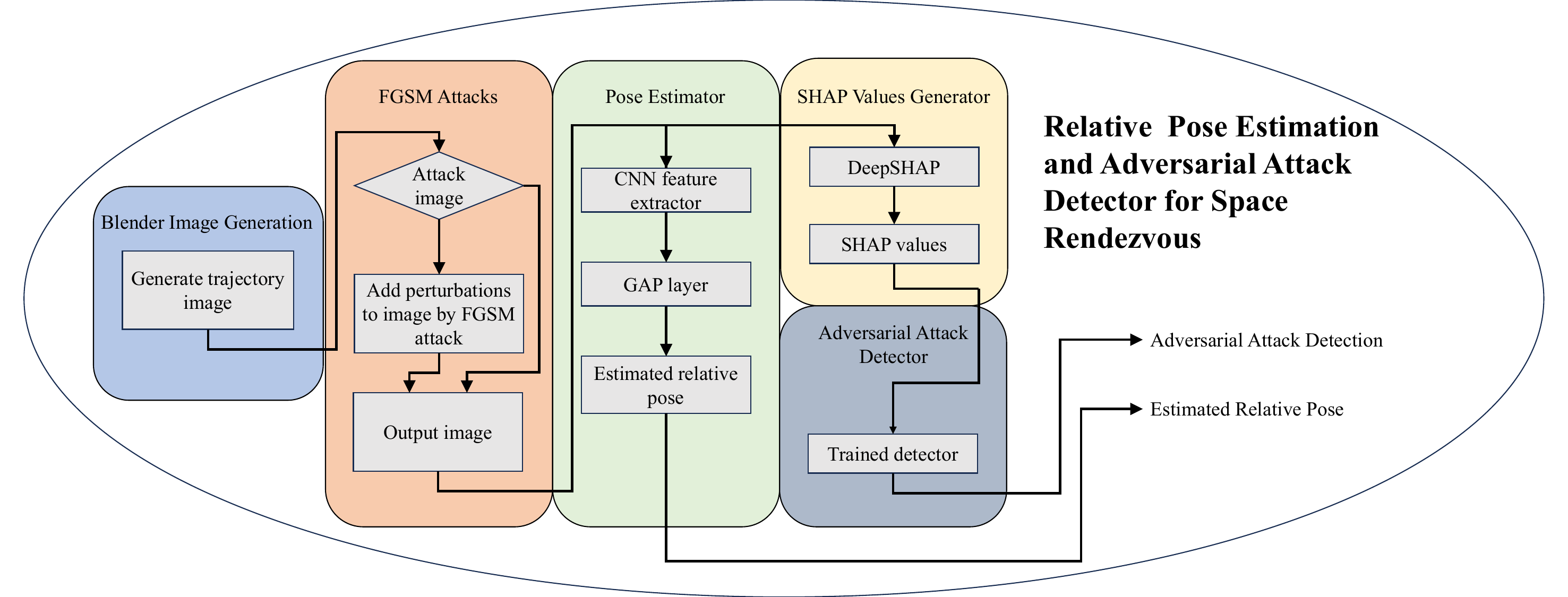}
    \caption{The experimental system includes the integration of an adversarial attack detector with the relative pose estimator and SHAP values generator.}
    \label{fig:integrated}
\end{figure*}

The SHAP values for both normal and adversarial samples are split into a training and testing set using a 0.8 train-test ratio, resulting in 24,000 samples for training and 6,000 samples for testing. The adversarial attack detector is trained using the Stochastic Gradient Descent (SGD) method with the Adadetal optimiser for 1,000 epochs. After training the adversarial attack detector, it achieved a training accuracy of 99.98\% and a test accuracy of 99.90\% on the test dataset. In this case, the detection accuracy is calculated by~\eqref{detect acc} 
\begin{equation}\label{detect acc}
    accuracy = \frac{successful~Detection}{Total~No.~of~Frames} \times 100\%
\end{equation}
where the $successful~Detection$ is defined by that the input frames with the adversarial attack are detector as $True$ and frames without adversarial attack are detector as $False$. The experimental results show that the proposed detector can successfully detect adversarial attacks on the DL-based relative pose estimator with high accuracy. The adversarial attack detector is integrated with the deep relative pose estimator and the DeepSHAP explainer to enhance accuracy in space rendezvous scenarios. The overall system is presented in Fig.~\ref{fig:integrated}. 

\begin{table}[htbp]
\centering
\caption{The average accuracy of the adversarial attack detector in test trajectories with various $\epsilon$ values.}
\begin{tabular}{ccc}

\hline
$\epsilon$            & Trajectory    & Detection   Accuracy \\ 
\hline
\multirow{3}{*}{0.5}  & 1             & 100\%                \\
                      & 2             & 100\%                \\
                      & 3             & 100\%                \\ 
\hline
\multirow{3}{*}{0.3}  & 1             & 100\%                \\
                      & 2             & 100\%                \\
                      & 3             & 99.98\%                \\ 
\hline
\multirow{3}{*}{0.1}  & 1             & 99.96\%                \\
                      & 2             & 99.98\%                \\
                      & 3             & 99.96\%                \\ 
\hline
\multirow{3}{*}{0.05} & 1             & 100\%                \\
                      & 2             & 99.98\%                \\
                      & 3             & 99.98\%                \\ 
\hline
\multirow{3}{*}{0.01} & 1             & 97.06\%                \\
                      & 2             & 96.94\%                \\
                      & 3             & 99.02\%                \\ 
\hline
Average               &               &99.21\%                   \\
\hline
\end{tabular}
\label{tab:synthetic integrated}
\end{table}

The adversarial attack detector is then tested with three trajectories. In each trajectory, the camera (spacecraft) starts 60 meters away from the target, positioned at various points in the $x$ and $y$ directions within the range of $[\pm 25, \pm 15]$. The camera is oriented directly toward the target, with an attitude represented as quaternion $(1,0,0,0)$. The end position is $(0,0,10)$. The camera (spacecraft) moves linearly at a rate of 0.25 meters per frame along the $z$-axis. It follows a projectile trajectory in the $x$ and $y$ directions, resulting in a total of 2,500 frames for each trajectory. The FGSM attack is applied to test trajectories with an attack probability of 0.2. Once FGSM is initiated, attacks continue for the subsequent 5 frames. The results of the proposed adversarial attack detector are presented in Table~\ref{tab:synthetic integrated}. From the test results, the proposed adversarial attack detector successfully detects all incoming FGSM attacks when the $\epsilon$ = 0.5. As the $\epsilon$ value goes small, i.e. fewer perturbations are made to input images, the detection accuracy has slightly dropped. For these three test trajectories, the proposed adversarial attack detector achieves a detection accuracy of 99.21\% on average. 

\subsection{Experimental Results on Real Data}

In previous experiments, both the proposed deep relative pose estimator and the adversarial attack detector exhibited high accuracy on synthetic data. To further evaluate the performance of both systems, we tested them with real-world images obtained from the Autonomous Systems and Machine Intelligence Laboratory (ASMI Lab) at City, University of London. These data include sensor noise, camera calibration noise, ground truth measurement noise, and different lighting conditions that are not present in the training synthetic images.

\subsubsection{Accuracy of the Spacecraft Dee Relative Pose Estimator}
\begin{figure}[h!]
    \centering
    \includegraphics[width=0.8\linewidth]{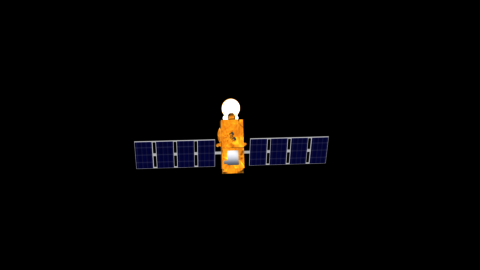}
    \caption{Am example of images generated from Blender for training the pose estimator.}
    \label{fig:real train}
\end{figure}

At the ASMI Lab, a scaled mock-up model of the Jason-1 spacecraft is constructed. This mock-up model is 1/9 the size of the actual Jason-1 spacecraft. The vision sensor applied for real data acquisition is the ZED 2 camera, which outputs images with a resolution of 1920$\times$1080. The deep relative pose estimator is retrained on new synthetic data, referred to as the Synthetic-Lab Dataset, with an input image size of 480$\times$270 to match the aspect ratio of the camera used in the ASMI Lab. As before, the Synthetic-Lab Dataset is generated using Blender, where the target was replaced with a 3-D model of the ASMI Lab mock-up Jason-1. To simulate the space rendezvous scenario over a distance range from 60m to 10m, the 3-D model is scaled up by a factor of 9 in Blender data generation. An example of the re-training images is shown in Fig.~\ref{fig:real train}

Similar to the previous synthetic data experiment, multiple trajectories are generated to collect images from the Blender, resulting in a total of 32,500 images on Synthetic-Lab Dataset for training and testing. The hyperparameter settings for training are the same as the settings applied in previous synthetic data experiment, including the learning rate, optimiser, batch size, and data augmentation methods. The pose estimator was trained for 100 epochs with a train/test split of 0.8. 

There are three sets of images captured from the ASMI Lab, referred to as the ASMI Dataset, with each set containing a total of 750 images. To acquire images for the ASMI Dataset, the camera movement is controlled by the Rethink Robotics Sawyer~\cite{Sawyer} moving along the $z$-axis, and the ground truths relative poses of the images in ASMI Dataset are recorded by the OptiTrack Motion Capture Systems~\cite{optitrack}. The OptiTrack Motion Capture System records the position and attitude of the ASMI Lab mock-up Jason-1 and the ZED camera at a frame rate of 120 frames per second and assigns a timestamp to each frame. Images are acquired by the ZED camera at a resolution of 1920 $\times$ 1080 and a frame rate of 30 frames per second, with relevant timestamps. The ground truth pose for each frame acquired by the ZED camera are assigned by matching the corresponding timestamps from the OptiTrack Motion Capture System. Then, the relative ground truth position is calculated by the difference between the actual positions of the ZED camera and the ASMI Lab mock-up Jason-1, as shown in equation~\eqref{eq:relative_pos_lab},
\begin{equation}\label{eq:relative_pos_lab}
    Pos_{lab} = Pos_{camera} - Pos_{target} 
\end{equation}
where $Pos_{lab}$ donates relative ground truth position in ASMI Dataset. The $Pos_{camera}$ and $Pos_{target}$ donate the actual position of the ZED camera and ASMI Lab mock-up Jason-1 recorded by OptiTrack Motion Capture System, respectively. 

\begin{figure}[h!]
  \centering

  \subfigure[]{%
    \includegraphics[width=0.48\linewidth]{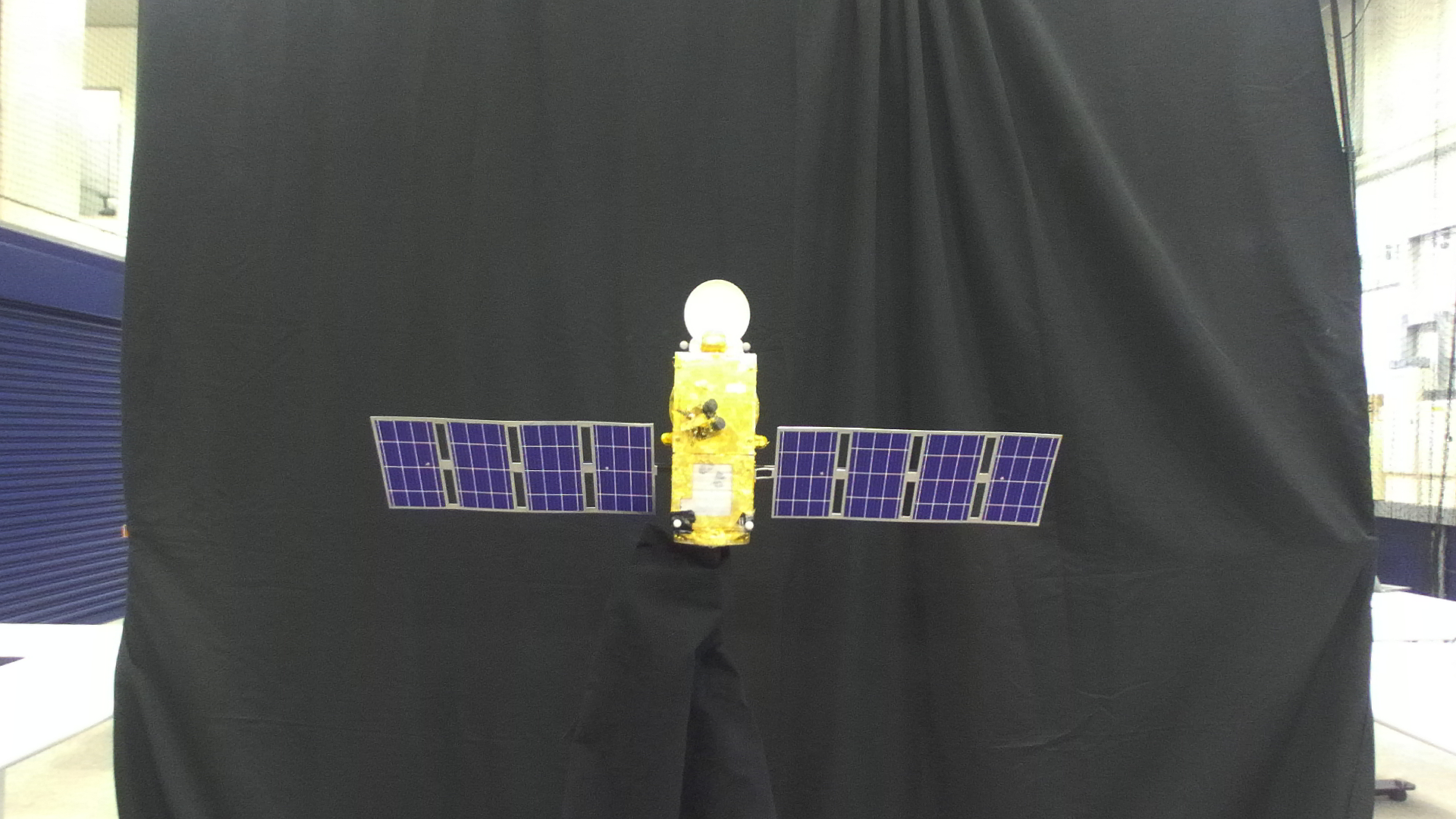}
    \label{fig:lab_ori}
  }\hfill
  \subfigure[]{%
    \includegraphics[width=0.48\linewidth]{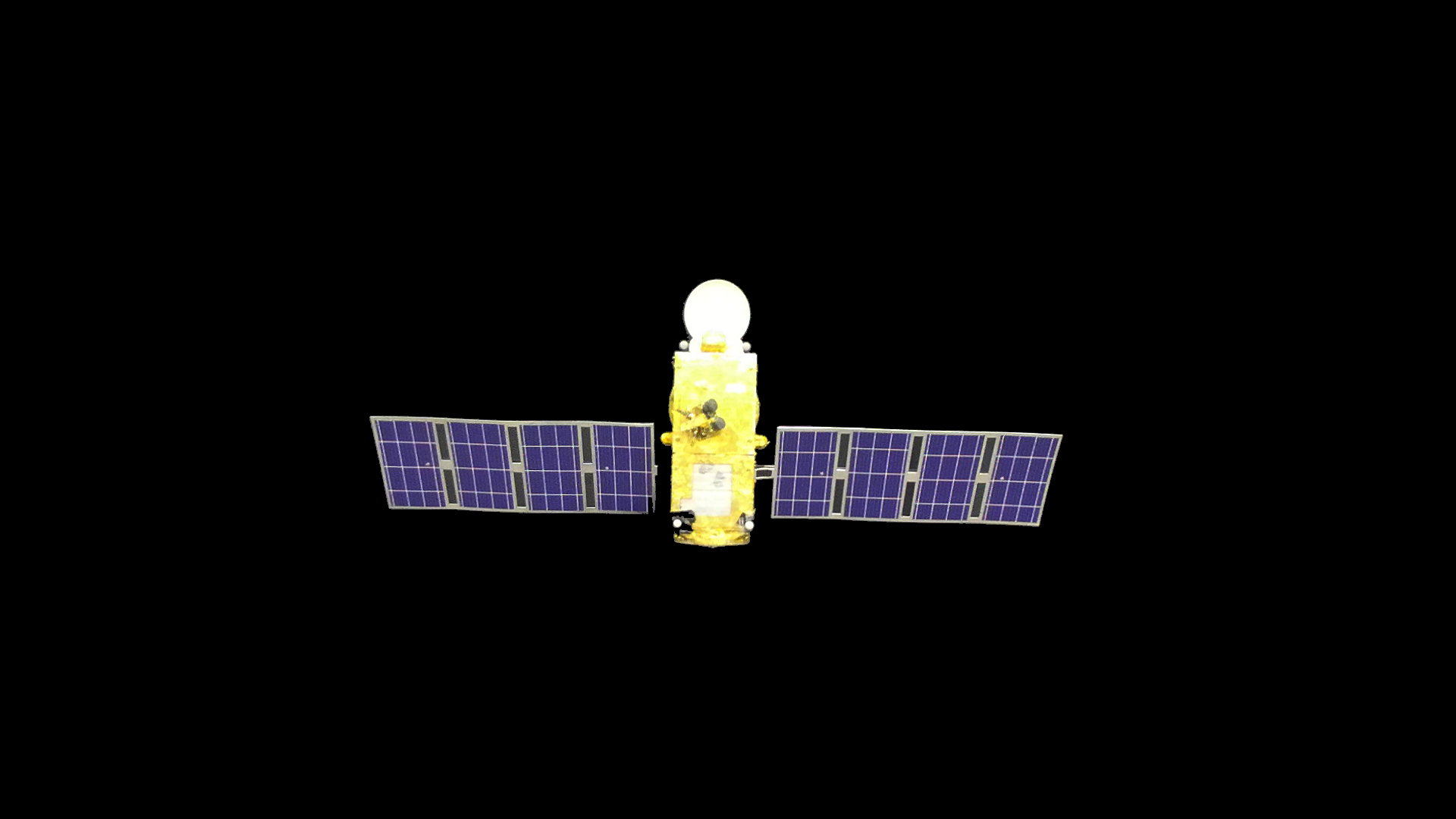}
    \label{fig:lab_seg}
  }

  \caption{Examples of images captured in ASMI Dataset. (a) Original image captured in ASMI Lab (b) Segmented image with bakc background.}
  \label{fig:real example.}
\end{figure}

Due to different camera intrinsic matrices applied between the Synthetic-Lab Dataset and ASMI Dataset, to represent the relative position in the trained model, the  position ground truths of the ASMI Dataset are collaborated with the camera view by the following processing:
\begin{equation}
    K_{Blender} =  \begin{bmatrix}
                640 & 0 & 240 \\
                0 & 360 & 135 \\
                0 &  0 & 1  \\
                \end{bmatrix}
\end{equation}
\begin{equation}
    K_{zed} =  \begin{bmatrix}
                1400.41 & 0 & 956.29 \\
                0 & 1400.41 & 557.258 \\
                0 &  0 & 1  \\
                \end{bmatrix}
\end{equation}
\begin{equation}
    Target_{real} = 9 \times Target_{lab} 
\end{equation}
\begin{equation}\label{pos}
    Pos_{real} = 1400.41 \times \frac{240}{956.29} \times \frac{Target_{lab}}{Target_{real}} \times \frac{1}{640} \times Pos_{lab}
\end{equation}
where $K_{Blender}$ and $K_{zed}$ represent the camera intrinsic matrices for the camera used in Synthetic-Lab Dataset collection and the ZED camera that is used to acquire images in the ASMI Lab, respectively. $Target_{real}$ and $Target_{lab}$ indicate the target sizes in the Blender 3-D model and the actual size in the ASMI Lab. $Pos_{real}$ and $Pos_{lab}$ denote the relative position of the target in the pose estimator and the ground truth position in the ASMI Lab, respectively. Table~\ref{tab:representative lab} illustrates the range of relative positions in the ASMI Dataset and representative relative positions in trained pose estimator. Furthermore, all images in the ASMI Dataset are segmented with a black background and resized to 480$\times$270 to fit the input image size of the trained pose estimator. An example of images captured in ASMI Lab is shown in Fig.~\ref{fig:real example.}.

\begin{table}[h!]
\centering
\caption{Camera moving range on ASMI Dataset and its representative range on trained pose estimator. The camera is moving along the $z-axis$. The representative range is calculated by equation~\eqref{pos}. }
\resizebox{\linewidth}{!}{\begin{tabular}{ccc}
\hline
trajectory ID & ASMI Lab Range (z-axis) & Representative Range(m) \\ 
\hline
ASMI-1             & 3.122 - 2.569           & 51.180 - 42.11        \\
ASMI-2             & 2.296 - 1.748           & 37.64 - 28.66         \\
ASMI-3             & 1.564 - 1.015           & 25.64 - 16.64         \\ 
\hline
\end{tabular}}
\label{tab:representative lab}
\end{table}

\begin{figure}[h!]
    \centering
    \includegraphics[width=\linewidth]{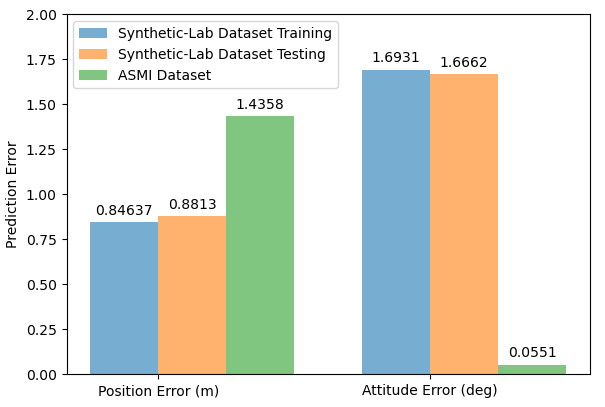}
    \caption{The prediction accuracy of the proposed pose estimator on Synthetic-Lab Dataset and ASMI Dataset after 100 epochs.The blue bar presents the average error on training data and the orange bar represents the average error on test data on Synthetic-Lab Dataset. The green bar indicates the average error on the ASMI Dataset.}
    \label{fig:real results}
\end{figure}

Once the deep relative pose estimation model is trained, it is initially tested on the test set of Synthetic-Lab Dataset, followed then by testing its prediction accuracy on real world data captured from the ASMI Lab, i.e. ASMI Dataset. The prediction accuracy of the deep relative pose estimator is reported in Fig.~\ref{fig:real results}. Similar to the previous synthetic testing, position error and attitude error are calculated by~\eqref{eq: pos err} and~\eqref{eq: rot err}, respectively. Compared with the prediction accuracy on the Synthetic-Lab Dataset, the position error of the ASMI Dataset is slightly higher. This could be attributed to variations in the illumination conditions compared to the Synthetic-Lab Dataset, as well as factors such as ground truth measurement noise and camera calibration noise. On the other hand, the predicted attitude error in the ASMI Dataset is much smaller than the synthetic data. One possible reason could be that the target remains stable at a fixed position with rotation effects during the images capture.

\subsubsection{FGSM Attacks on ASMI Dataset}
\begin{figure}[h!]
    \centering
    \includegraphics[width=\linewidth]{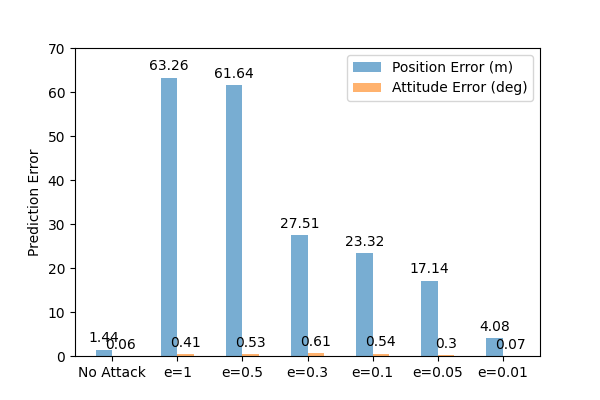}
    \caption{Comparison of the prediction error under FGSM attack on ASMI Dataset with various $\epsilon$ values. The blue bar indicates the average position error in meter and the orange bar represents the average attitude error in degrees. }
    \label{fig:real attack}
\end{figure}

To evaluate how the pose estimator can be impacted by adversarial attacks on real data, the FGMS attack is then applied to the ASMI Dataset. In this case, the FGSM is configured as the same $\epsilon$ as previously applied in synthetic, including 1, 0.5, 0.3, 0.1, 0.05 and 0.01. In this experiment, all images are perturbed by the FGSM attack. The model's average prediction error under FGSM attacks with various $\epsilon$ values on the ASMI Dataset are illustrated in Fig.~\ref{fig:real attack}.

As shown in Fig.~\ref{fig:real attack}, FGSM has a significant impact on position estimation but only slight impacts on attitude estimation. In comparison to the previous experiment with synthetic data, the FGSM attack has a more pronounced effect when $\epsilon$ is less than 0.05 on the predicted position in the ASMI Dataset. However, the attitude error remains quite stable, typically less than 1 degree, for all tested $\epsilon$ values. 

\subsubsection{LSTM-based Adversarial Attack Detector}

To evaluate the adversarial attack detector on the ASMI Dataset, SHAP values are obtained by processing the pose estimator on the Synthetic-Lab Dataset. Similar to the previous synthetic data experiment, the SHAP values are obtained from the output of the GAP layer in the deep relative pose estimator by DeepSHAP algorithm. 1,000 images from the training data on Synthetic-Lab Dataset are randomly selected to generate background data. A total of 30,000 SHAP value samples, consisting of 15,000 normal samples and 15,000 adversarial samples, are used to train the adversarial attack detector. The 15,000 normal samples consist of all images from the test data on the Synthetic-Lab Dataset and randomly selected images from the training data to account for the difference between 15,000 and the total number of images in the test data. Adversarial samples are generated by applying FGSM attacks to the normal sample images with randomly selected $\epsilon$ values from [0.5, 0.3, 0.1, 0.05, 0.01]. 

The SHAP values are shuffled and split by a train-test ratio of 0.8, i.e. 24,000 samples for training and 6,000 samples for testing. The adversarial attack detector is trained by SGD with an Adadelta optimiser for 2000 epochs. Early termination is implemented to reduce the training time. To do that, the training data are further split into 80\% for training and 20\% for validation. If the validation loss does not decrease over 20 epochs, the training process will be terminated. After the early termination condition, the proposed adversarial attack detector achieves a detection accuracy of 99.18\% on training data and 98.8\% on test data.

Subsequently, the pose estimator, FGSM attacks, and adversarial attack detector are integrated to evaluate the overall performance on the ASMI Dataset. The integrated system is identical to the one shown in Fig.~\ref{fig:integrated}, with the exception that the 'Blender Image Generation' part is replaced by the ASMI Dataset. In the ASMI Dataset, a random attack probability of 0.2 is applied to FGSM attacks. When an attack occurs, input images are continuously perturbed by FGSM for the next 10 frames. The detection accuracy is calculated by~\eqref{detect acc}. Table~\ref{tab:real integrated} presents the detection accuracy on the ASMI Dataset for various $\epsilon$ values.

\begin{table}[htbp]
\centering
\caption{The average accuracy of the adversarial attack detector in ASMI Dataset with various $\epsilon$ values.}
\begin{tabular}{ccc}

\hline
$\epsilon$      & Detection   Accuracy  \\ 
\hline
0.5             & 100 \%                \\
0.3             & 100 \%                \\
0.1             & 100 \%                \\
0.05            & 98.44\%               \\
0.01            & 90.44 \%              \\
\hline
Average         & 96.29 \%              \\
\hline
\end{tabular}                   
\label{tab:real integrated}
\end{table}

As shown in Table~\ref{tab:real integrated}, the proposed adversarial attack detector achieves an average correct detection rate of 96.29\% on the ASMI Dataset. The accuracy slightly drops when the $\epsilon$ value becomes smaller, which is caused by fewer perturbations applied to the input images as $\epsilon$ decreases.

\section{Conclusion} \label{conclusion}

This paper firstly examines the impact of adversarial attacks on DL-based spacecraft relative pose estimation in space rendezvous scenarios. To do this, a CNN-based relative pose estimation algorithm is proposed. FGSM adversarial attacks are implemented, which have a significant impact on the model's predictions. Subsequently, an LSTM-based adversarial attack detector is proposed to identify adversarial attacks on input images. XAI techniques are adopted to analyse the model's predictions and generate SHAP values-based explanations for the model's predictions. Multiple experiments are carried out to evaluate the performance of the CNN-based spacecraft relative pose estimator, how the adversarial attacks can impact on DL-based pose estimator in space rendezvous missions, and the performance of the proposed adversarial attack detector. The proposed methods have been tested on both synthetic and real image datasets. The results show that the adversarial attack detector performs robustly in detecting adversarial attacks, achieving an average of 99.21\% detection rate on synthetic data and 96.29\% on real data collected from the ASMI Lab. 

Although the impact of digital adversarial attacks on DL-based spacecraft relative pose estimation has been analysed in this work, how to physically implement the adversarial attacks still needs to be explored. Moreover, the proposed method demonstrates high accuracy in detecting adversarial attacks for the DL-based spacecraft relative pose estimation, how to correct the estimated pose after detecting adversarial attacks becomes critical to provide a robust DL-based system for future space missions. 

\section*{Acknowledgement}
This work is funded by the European Space Agency (ESA), Contract Number: ESA AO/2-1856/22/NL/GLC/ov.

\printcredits

\bibliographystyle{model1-num-names}
%\bibliographystyle{cas-model2-names}

\bibliography{reference}

\end{document}